\documentclass[openany]{ecai}
\usepackage{graphicx}
\usepackage{multirow}
\usepackage{latexsym}
\usepackage[title]{appendix}
\usepackage{tikz}
\usepackage{xspace}
\usepackage{caption}
\usepackage{forest}
\usepackage{subcaption}
\usepackage{url}

\usepackage{dirtree}

\newcommand{\cfxai}{cf-\textsc{XAI}\xspace} 
\newcommand{\nxai}{n-\textsc{XAI$^T$}\xspace} 
\newcommand{\basexai}{n-\textsc{XAI$^B$}\xspace} 
\newcommand{\bxai}{n-\textsc{XAI$^{B1}$}\xspace} 
\newcommand{\bbxai}{n-\textsc{XAI$^{B2}$}\xspace} 

\usepackage[absolute]{textpos}
\begin{document}

\begin{textblock}{12}(8,1)  
  \begin{minipage}{12cm}  
    \vspace*{-0.85cm} 
    \raggedleft  
    \textit{ECAI 2023 K. Gal et al. (Eds.)} \\
    \textcopyright{} \textit{2023 The Authors.} \\
    \textit{This article is published online with Open Access by IOS Press and distributed under the terms} \\ \textit{of the Creative Commons Attribution Non-Commercial License 4.0 (CC BY-NC 4.0).} \\
    \textit{Pages 2057 - 2064} \\
    \textit{DOI 10.3233/FAIA230499}
  \end{minipage}
\end{textblock}

 \begin{frontmatter}

 \title{Towards Feasible Counterfactual Explanations: A Taxonomy Guided Template-based NLG Method}

\author[A]{\fnms{Pedram}~\snm{Salimi}}
\author[A]{\fnms{Nirmalie}~\snm{Wiratunga}}\orcid{0000-0003-4040-2496}
\author[A]{\fnms{David}~\snm{Corsar}}\orcid{0000-0001-7059-4594}
\author[A]{\fnms{Anjana}~\snm{Wijekoon}}\orcid{0000-0003-3848-3100}

\address[A]{Robert Gordon University, Aberdeen, UK}

 \begin{abstract}
 Counterfactual Explanations (\cfxai) describe the smallest changes in feature values necessary to change an outcome from one class to another. 
 However, many \cfxai methods neglect the feasibility of those changes. 
 In this paper, we introduce a novel approach for presenting \cfxai in natural language (Natural-XAI), giving careful consideration to actionable and comprehensible aspects while remaining cognizant of immutability and ethical concerns.
 We present three contributions to this endeavor. Firstly, through a user study, we identify two types of themes present in \cfxai composed by humans: content-related, focusing on how features and their values are included from both the counterfactual and the query perspectives; and structure-related, focusing on the structure and terminology used for describing necessary value changes.
 Secondly, we introduce a feature actionability taxonomy with four clearly defined categories, 
 to streamline the explanation presentation process.
 Using insights from the user study and our taxonomy, we created a generalisable template-based natural language generation (NLG) method compatible with existing explainers like DICE, NICE, and DisCERN, to produce counterfactuals that address the aforementioned limitations of existing approaches.
 Finally, we conducted a second user study to assess the performance of our taxonomy-guided NLG templates on three domains.
Our findings show that the taxonomy-guided Natural-XAI approach (\nxai) received higher user ratings across all dimensions,  with significantly improved results in the majority of the domains assessed for articulation, acceptability, feasibility, and sensitivity dimensions.

 \end{abstract}

 \end{frontmatter}

 \section{Introduction}
\label{introduction}

A counterfactual explanation (\cfxai) shows how to get a different outcome from a black-box AI model by changing only a few input features.
This aligns with human intuition by offering the black-box model's underlying rationale in the form of a counter-argument~\cite{ruth-psyc}.
It serves three primary goals~\cite{wachter2017counterfactual}:
1) elucidate the reasoning behind decisions;
2) supply adequate information to critique decisions with negative outcomes; and
3) enable a better understanding of the necessary changes to achieve desired outcomes in the future.
There is an abundance of techniques to generate \cfxai in the literature that achieve some subsets of these three goals~\cite{nice, guidotti2022counterfactual, dice, verma2020counterfactual, discern}. The focus of this paper instead is to achieve the third goal as a post-processing step taking into account the user perspective. 

The literature identifies many properties of good counterfactuals, such as sparsity, proximity, validity, diversity, feasibility, and plausibility~\cite{keane}. 
To achieve the third goal, feasibility, must be integrated into the explanation generation process, such that the resulting \cfxai
provides a complete understanding of suggested changes. 
The challenge in attaining feasibility lies in meticulously evaluating the suggested alterations in the context of the user, taking into account factors such as appropriateness and ethics. For instance, proposing a change in an individual's weight might be acceptable in the medical domain; however, within the social sphere, such recommendations could be perceived as disrespectful and potentially offensive.
The question of how to  
guide the user through the recommended changes while being sensitive to the types of features when presenting a \cfxai remains unanswered.

To the best of our knowledge, no formal approach 
exists for managing user-specific feasibility considerations when presenting the counterfactual's recommended changes. 
Certain methods~\cite{dice} generate a diverse set of counterfactuals in the expectation that users will identify one or more as feasible; others~\cite{nice} entrust individuals with specifying requirements that can be integrated into the counterfactual generation algorithm. 
Both impose considerable cognitive load on the user.
In this paper, we formalise feasibility requirements using a taxonomy to enable a natural language presentation of explanations (Natural-XAI),
using a template-based natural language generation (NLG) approach to effectively address and handle feasibility-related criteria when presenting counterfactual recommendations.
\noindent Accordingly we make the following contributions:
\begin{itemize}
\item propose a set of common natural language constructs, identified from a user study, that enables us to convey \cfxai in a better textual presentation format;
\item introduce a taxonomy that captures the knowledge of \textit{feature actionability} and categorises features based on their mutability;
\item present a template-based NLG method (\nxai) that utilises the feature actionability taxonomy to generate counterfactual Natural-XAI;
\item conduct a user study analysis of the proposed \nxai method across three application domains, demonstrating that it improves counterfactual understandability with respect to sensitivity, acceptability, feasibility, and articulation; and
\item provide useful guidelines and insights for XAI platform development, derived from a thematic analysis of user responses.
\end{itemize}

In Section~\ref{relatedworks} we present related work, while Section~\ref{sec:userstudy} describes an initial user study conducted to gather insights for improving counterfactual Natural-XAI.
Section~\ref{sec:taxonomy} formalises actionable recommendations for \cfxai systems, based on our proposed taxonomy of feature actionability, and Section~\ref{sec:nlg} describes a mapping from taxonomic categories to language generation templates for Natural-XAI.
This section also discusses in detail our proposed \nxai method, and the effectiveness of \nxai, which incorporates a three-stage NLG pipeline integrating actionability knowledge, and evaluated in a second user study with results presented in Section~\ref{sec:userstudy2}.

 \section{Related Work}
\label{relatedworks} 

A \cfxai is distinct from a factual explanation, as it aims to answer "What-If" and "Why-Not" user queries that relate to the input-output relationship of a black-box model, while factual explanations typically address "Why" questions.
Given a user query and the AI's prediction, a \cfxai defines the smallest change in feature values required to shift a prediction to the desired outcome. 
For example, in response to an AI loan application system's prediction, a \cfxai may propose, "A smaller loan amount would have resulted in your application being accepted", where the action of decreasing by a small amount is the proposed action recommended by the \cfxai system. 

Proponents of counterfactual theories argue that they offer significant computational, psychological, and legal benefits~\cite{keane}.
Effective and interpretable \cfxai must also satisfy several key requirements. Sparsity calls for minimising the number of modified features, while proximity ensures that the counterfactual instance is as close as possible to the original instance in the feature space, thereby seeking the minimal change necessary to achieve the desired outcome~\cite{keane}. 
Both can be addressed either by case-based instance learning~\cite{nice,craw_cbr,discern} or as parameters within optimisation minimisation techniques~\cite{dice,wachter2017counterfactual}.
Feasibility ensures suggested changes are achievable~\cite{poyiadzi2020face}, and plausibility maintains realistic distributions~\cite{yang2020generating}.
This paper concentrates on feasibility and recourse, which involve users taking actionable steps based on provided explanations to achieve desired outcomes~\cite{ustun2019actionable}.
Feasibility, refers to whether a proposed change can realistically occur, and actionability, concerns the user's capacity to implement the change.
Recourse emphasises the importance of suggesting feasible changes that users can implement realistically, to maintain user trust in AI systems.
Specifically, we investigate post-processing generated explanations to refine their presentation format while carefully considering the feasibility of suggested actions, enabling our method to integrate with any \cfxai algorithm.

Numerous studies have explored feasibility of counterfactuals, such as the FACE algorithm~\cite{poyiadzi2020face}, which generates feasible counterfactuals by considering proximity in high-density regions. However, generalising to all individuals may be challenging due to diverse backgrounds and situations, rendering certain feasible counterfactuals ineffective for some users~\cite{barocas2020hidden}.
While feasibility knowledge aids in post-processing generated explanations by guiding the selection of suitable presentation styles for each recommended action, causal knowledge helps with grouping interrelated actions and effectively presenting actionable groups.
However, most counterfactual explainers do not consider causal relationships. 
For example, while ~\cite{dice} highlights the importance of causal constraints for feasibility, they do not offer methods for generating them. 
Similarly,~\cite{karimi2021algorithmic} emphasises causality and human intervention in feasible \cfxai, but their approach necessitates deep causal model understanding. 
Although ~\cite{mahajan2019preserving} presents a variational autoencoder-based learning method for generating feasible counterfactuals, it lacks scalability across domains and adequate data for learning causal constraints.
Here, we focus on feasibility aspects in post-processing \cfxai outputs, and argue that if causal knowledge is available, the presentation would involve combining presentations of individual features, which is less challenging compared to addressing the presentation of feasibility aspects.

Natural-XAI is more human-friendly and can be tailored to the user’s specific context, beliefs, and preferences~\cite{nile}. 
For instance, textual data representation has been shown to outperform visual graphs in clinical decision-making
~\cite{law2005comparison,petre1995looking}, enhancing trust, transparency, acceptability, and usability.
Effectively presenting counterfactuals requires managing actionable changes, which can be difficult to absorb when presented in tabular form. 
A natural language format is likely to be more accessible
and capable of clearly describing the recommended actionable changes. 
In domains such as finance and health, controlled text generation, an advancement in numerical reasoning for language models, is critical for accurately conveying \cfxai where inaccurate suggestions for numerical values or attributes can have significant consequences, such as rejected loan applications and financial harm to the applicant~\cite{suadaa2021towards}.
Despite advancements in large language models like GPT-3, such inaccuracies persist~\cite{ji2022survey} due to hallucinations in generated text. 
The alternative template-based approach 
is a more reliable solution to integrate feasibility and ethical considerations for Natural-XAI.
In this paper, we use a taxonomy to generate feature-based templates that inform the NLG process in Natural-XAI sentence planning,
with surface realisation focused on choosing comparative adjectives, action verbs, and other forms of language constructs to convey actionable changes based on the taxonomy node type.
Discourse planning involving the ordering of these sentences is typically influenced by the importance of the recommended action, based on feature attribution explainer weights~\cite{lundberg2017unified}.
 \section{Understanding How to Compose Counterfactuals}
\label{sec:userstudy}
To understand how counterfactuals are authored, we carried out a user study examining naturally expressed counterfactuals in the widely-used loan-approval dataset, with the aim of identifying  reusable linguistic constructs without requiring domain expertise.

\subsection{User Study Setup}

\begin{figure}[t]
    \centering
    \includegraphics[width=.5\textwidth]{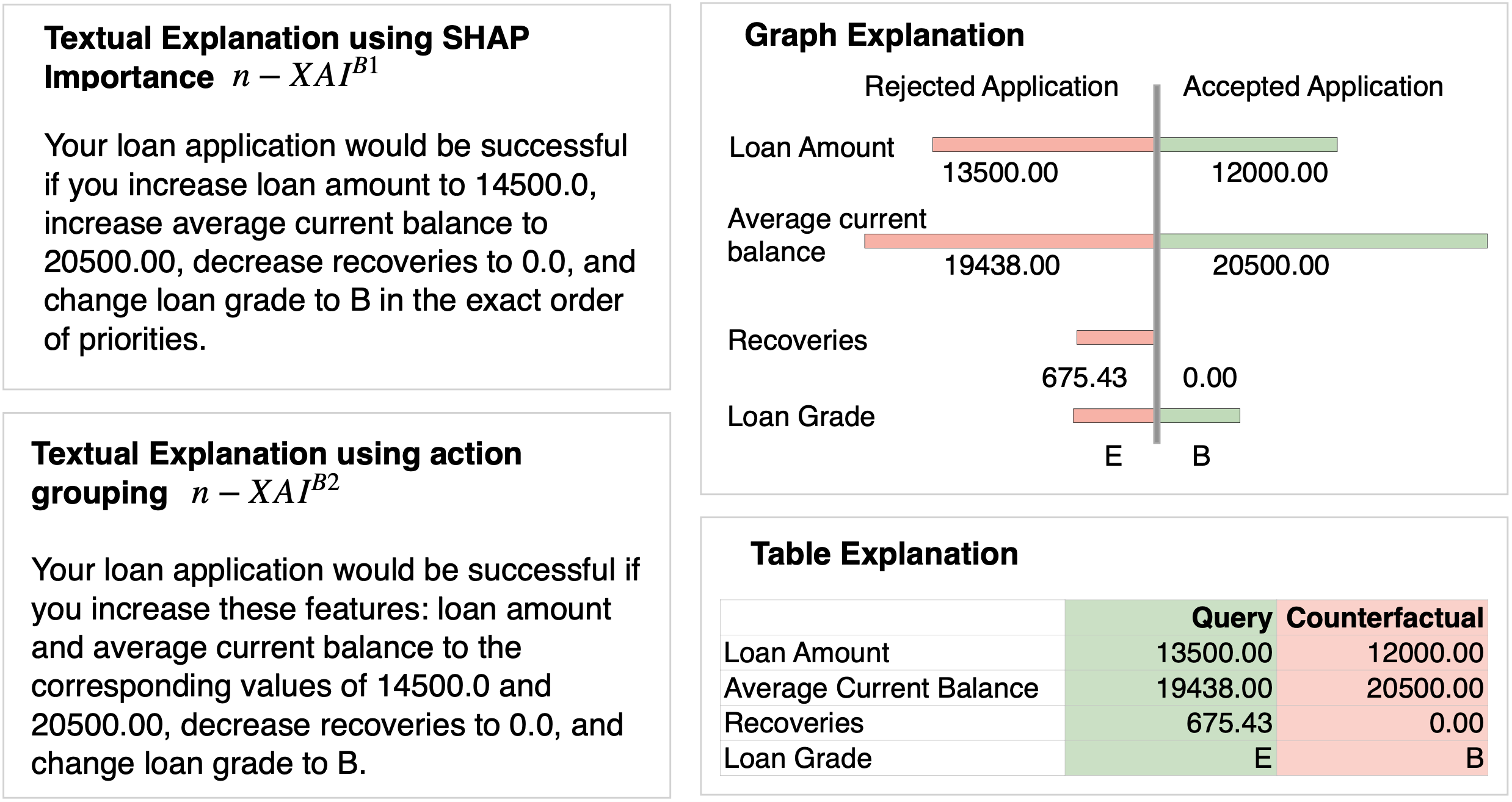}
    \caption{Examples of \cfxai presentation formats}
    \label{fig:cf-formats}
\end{figure}

Using DICE~\cite{dice}, counterfactuals with 4-5 actionable feature changes for seven \textit{rejected} loan applications was selected.
Here the choice in the number of changes was based on the \cfxai's actionable changes distribution observed on the loan dataset. 
A query to the \cfxai system provides feature value pairs, and the corresponding counterfactual recommends alternative values for a subset of these features to achieve a desired outcome from the black-box model.
For each such \{query \& counterfactual\} pair, we generate a tabular form of the counterfactual that includes recommended changes to the query feature values ordered by SHAP~\cite{lundberg2017unified} local feature importances. 
Additionally, we create a zero-centered visual chart (maintaining SHAP orderings) and a basic Natural-XAI format as alternative presentations of the same counterfactual 
(see Figure~\ref{fig:cf-formats}). 
For the Natural-XAI presentation, 
two basic NLG templates were used: 
\bxai, where feature changes are presented in order of SHAP feature value importance (e.g., increase feature F1 to V1 then decrease feature F2 to V2 $\ldots$); or 
\bbxai, where feature changes are ordered by SHAP but grouped by action (verb) type order 
(e.g., increase features F1 \& F3 to values V1 \& V3, thereafter decrease features \ldots).

We had two independent cohorts, with alternative explanation formats allocated as follows: 
Cohort1 was shown a tabular form of the counterfactual and a visual chart highlighting differences between query and counterfactual, while Cohort2 was presented with the Natural-XAI forms. 
In setting up the study we wanted to:
1) minimise the impact on users' natural expressions of counterfactuals by presenting multiple alternatives to address potential biases, such as cognitive, visual, and familiarity biases;
2) examine if access to basic natural XAI would influence the quality of users' responses;
3) determine if any correlation exists between the authored text and either \bxai or \bbxai concerning preferences for sentence ordering; and
4) assess the extent to which access to alternative formats might influence the quality of user responses.

\subsection{User Study Protocol}
The study was conducted in October 2022 with 33 participants aged 18-24 from an undergraduate AI course but prior to learning about XAI and with no prior domain knowledge, 
although it is reasonable to assume that some familiarity with "student loans" is to be expected amongst a majority of our participants.
After providing details about the dataset, including possible prediction outcomes and descriptions of loan features, participants were divided into cohorts. 
They received a query and corresponding counterfactual in alternative formats depending on their cohort.
Participants were asked to use these counterfactuals to complete the tasks listed below:
\begin{enumerate}
\item Task1: Compose a piece of natural language guidance for the loan applicant to help them achieve a better outcome in the future.
\item Task2: Evaluate and rank the alternative explanation presentations based on their usefulness for composing the recommendation in Task1, along with justifications for the preferences.
\item Repeat Tasks 1 and 2, for 4 query instances from a set of 7 randomly selected queries.
\end{enumerate}
In Task2, each cohort had access to different formats of the counterfactual: Cohort1 had table and chart; and Cohort2 had \bxai and \bbxai.
We expect that responses to Task1 will enable better understanding of the sentence planning requirements for Natural-XAI and provide insights into discourse planning, such as determining the most effective order in which to present content. 
Task2 aims to assess the degree to which the SHAP ordering of features in the table and chart influences the order of authored text for Cohort1. 
For Cohort2, the task seeks to identify any preferences or influences resulting from the ordering presented in both the SHAP-based \bxai and action group-based \bbxai. 
Accordingly, Task2 is expected to provide valuable insights for discourse planning.

\subsection{Quantitative Evaluation}
\begin{table}
\centering
\renewcommand{\arraystretch}{1.1}
\begin{tabular} {lp{2.6cm}rr}
\hline
Criteria & Analysis & Cohort1 & Cohort2\\ 
\hline
\multirow{4}{*}{\parbox{1.1cm}{Response Statistics}} & $\#$ of samples & 55       & 53 \\
& Min. Length    & 2        & 4 \\ 
& Max. Length    & 63      & 75 \\
& Ave. Length    & 22.86        & 31.35 \\
& Std.Dev.    & 14.48       & 14.29\\
\hline
\multirow{2}{*}{Readability} & Flesch Score &  52.79 & 41.93\\
& Style Description & Complex  & V.Complex\\

\hline
\multirow{4}{*}{Grammatical}& Mean Error &1.85&2.05\\
& Error-free frequency&16&18\\
& Max Errors           &8&7\\
\hline
\multirow{2}{*}{Avg Similarity}& Token-wise &\ 7\% &23\%\\
& Semantic &55.4\%&70.3\%\\
\hline
\multirow{2}{*}{\parbox{2cm}{Ordering Correlation Analysis}} & SHAP  & 17.28 & 34.75\\
 & Action group  & -- & 78.03 \\
\hline
\multirow{2}{*}{\parbox{1.9cm}{Alternatives Preferred }}& \bbxai $>$ \bxai & -- & 55\%\\
& Chart $>$ Tabular & 84.61\% & --\\
\hline
\end{tabular}
\caption{Response analysis by cohort}
\label{table:t1-responses}
\end{table}
The quantitative text analysis of the 108 responses used 7 criteria (see Table~\ref{table:t1-responses}). 
Here the "Alternatives Preferred" criteria relate to Task2 whilst the rest relate to Task1. 
\begin{description}
\item[Response Statistics] 
for 108 responses were analysed after excluding brief or non-compliant submissions. 
Length statistics were calculated without 4 max length outliers.
A minimum length was 2 to accommodate responses like "Higher total\_rec\_prncp"

\item[Readability] is assessed using the Flesch score~\cite{flesch1948new} which calculates a value between 0-100, 
where lower values indicate lower readability due to complex constructs. We find that Cohort2 participants responded with more complex explanations i.e. Style Description = Difficult, where text is likely to contain, longer sentences, technical vocabulary, and more specialised ideas. 

\item[Grammatical error analysis] was conducted using the Python LanguageTool (for spelling and grammar).
Cohort2 outperformed Cohort1 suggesting that basic Natural-XAI formats of \bxai and \bbxai had influenced the cohort to write more grammatically correct explanations, compared to Cohort1 who only saw tabular and graphical formats.
Token-wise average similarity between Cohort2's text compositions and \bxai and \bbxai showed a 23\% match, indicating that participants did not merely copy-paste the content.

\item[Semantic similarity] uses the all-MiniLM-L6-v2 model from the sentence-transformers library, to assess the semantic similarity between both cohorts' authored text to the \bxai and \bbxai Natural-XAI forms. 
As expected results showed an average resemblance of 55.4\% for Cohort1 and 70.28\% for Cohort2. 
Since \bxai and \bbxai were factually correct, a greater resemblance to these baselines can be used as a reliable measure of the plausibility of the participant-generated explanations.

\item[Correlation] between the feature ordering methods of \bxai and \bbxai, which are based on the SHAP and Action Group models, and the ordering of features in the text generated by Cohort2 were compared using the Spearman's rank order coefficient. The analysis indicated that Cohort2 preferred grouping their text by actionability verbs before ordering features by SHAP order. 
However, Cohort1 did not exhibit a significant correlation in the ordering of features, despite being presented with SHAP ordering through the tabular and graphical alternatives.
\item[Alternatives Preferred] relate to Task2, and findings suggest that 55\% of cohort 2 preferred \bbxai and 84.61\% of Cohort1 preferred chart over tabular explanation.
\end{description}

\subsection{Qualitative Evaluation - Thematic Analysis}
\begin{table*}[ht]
\centering
\footnotesize
\renewcommand{\arraystretch}{1.3}
\begin{tabular}{p{3cm}p{11.5cm}cp{1.0cm}}
\hline
Content Theme & Example & Frequency & Cluster(s) \\
\hline
C1:\textbf{Vague} action & 
The customer should \textbf{lower} their \textbf{collection fee} and their \textbf{recoveries} as well as \textbf{increase} their \textbf{total payment} and their \textbf{principal} & 23 & 1, 4, 5, 3 \\
C2: \textbf{Counterfactual} values only & 
For your application to be improved, you would need to \textbf{increase total\_pymnt to 12268.08} and \textbf{total\_rec\_prncp to 10925}.You would need to \textbf{decrease recoveries and collection\_recovery\_fee to 0} & 39 & 0, 2 \\
C3: \textbf{Counterfactual and Query} values & 
For a successful application you should increase total payment \textbf{to 12268.08 from 5040.00} while also increasing principal \textbf{to 10925.00 from 1492.93} and decrease recoveries and collection fee to 0 & 5 & 0, 2 \\
C4: \textbf{Combined} T2 \& T3 &
For your application to be successful you should increase your \textbf{principal to 14999.99} and your \textbf{fico to 669.00 from 585} while \textbf{decreasing recovery} and \textbf{collection recovery fees to 0} and \textbf{decreasing} your \textbf{last payment amount to 33.35}   & 3 & 0, 2 \\
C5: \textbf{Reduced} explanation & 
\textbf{Reduce} the \textbf{recoveries} to \textbf{0} and \textbf{total\_rec\_prncp} needs to be \textbf{higher} & 20 & 0, 5 \\
\hline
Structure Theme & Example & Frequency & Cluster(s) \\
\hline
S1. Use of ordinal adverbs / ordering with Bullet-pointing & 
For your application to be accepted, you will need to prioritize the following in the \textbf{order} given: \textbf{1)} \textbf{increase} the total principal received (total\_rec\_prncp) to 14,999 \textbf{2)} \textbf{decrease} the recoveries (recoveries) to 0.00 \textbf{3)} \textbf{decrease} the collection recovery fee (collection\_recovery\_fee) to 0.00 \textbf{4)} \textbf{increase} the last payment amount (last\_pymnt\_amnt) to 11,448.66 & 12 & 2 \\
S2. Creative Action Verbs & 
Cust1 should \textbf{pay off} their recoveries and \textbf{negotiate} to have their charge off \textbf{removed} & 4 & 2 \\
\hline
\end{tabular}
\caption{Content and structure themes with examples and alignment to response clusters.}
\label{tab:user-study-discussion}
\end{table*}

The qualitative evaluation consists of two steps: 
1) analyse all authored explanations manually to identify common themes; and 
2) conduct a clustering to compare automatically created clusters with the manually formed themes.
The manual analysis was conducted using a thematic analysis approach, where the content was coded by two researchers. Common themes were then aggregated and defined.
This process identified five content-related and two structure-related themes. 
Table~\ref{tab:user-study-discussion} lists the themes, with examples and the frequency of each theme in the responses. 
The "Content" theme identified five variations in the text authored by the study participants,
based on the presence or absence of feature values with reference to the query, counterfactual, or both.  
The "Vague" theme refers to responses where the recommended actions remained unclear (i.e. the exact amount by which to change was missing), 
while "Reduced" indicates responses where mentions of subsets of features and actions were missing. 
The "Structure" themes were examined using ordinal adverbs and ordering styles (e.g. bullet pointing), and the use of unusual/interesting "actionable words" (verbs).
Agglomerative clustering was used on responses to assess alignment with manually extracted themes.
The average linkage method is used to merge the clusters based on their similarity. The textual responses were encoded using the  pre-trained all-MiniLM-L6-v2 model from the sentence transformers library to create embeddings, which were used as input to the clustering algorithm. 
The resulting clusters were compared to the manually identified themes, and the similarity was assessed (see Table \ref{tab:user-study-discussion}). 
We found that the clustering results were mostly consistent with the manually identified themes, with the "Vague" theme split into many smaller clusters 
separate from the other themes.
Specifically, we found that themes 2, 3, and 4 were highly similar and mostly clustered together in clusters 0 and 2. 
These related to the differences as to whether or not feature names and their values were mentioned with reference to the counterfactual, query or both. 
For instance, consider these two alternative sentences to convey an actionable change:  
"s1: your loan amount of 13,5K needs to be reduced to 12K" and "s2: you must reduce your loan to 12K". 
Here s1 uses feature values from both the query and counterfactual whilst s2 only refers to the counterfactual. 
Results show that actionable changes are more frequently referenced with counterfactual values and feature names, rather than using query values.

\subsection{Findings for Counterfactual Natural-XAI}
Our analysis suggests that the use of Natural-XAI formats, such as \bxai and \bbxai, may have had a positive impact on the quality and accuracy of the explanations generated by users, as evidenced by the higher level of grammatical correctness observed in Cohort2 and the preference for organising text by actionability verbs before ordering features by SHAP order. 
Accordingly, to help with discourse planning we can adopt such ordering strategies to organise sentences. 
For sentence planning, employing action verbs like "negotiate" and phrases like "strive to" highlights the need for actionability concepts that can capture varying degrees of actionability.
The factual similarity and the variation in semantic similarity observed in Cohort2's text to \bxai and \bbxai suggest value in further studying Cohort2's content and structural organisation to derive generalisable Natural XAI templates. 

Our thematic analysis integrates findings from both content and structure analysis to inform two components. 
First, we develop a taxonomy in Section \ref{sec:taxonomy} that discerns different categories of feature actionability. 
Second, we create Natural-XAI templates in Section \ref{sec:nlg} that encapsulate common constructs based on ordinal adverbs, comparative adjectives, and action verbs. 
These templates are inspired by and take ideas from the human-authored text in the user study.

 \section{Counterfactual Natural-XAI Method}
Our Natural-XAI template-based NLG method, \nxai, uses a Feature Actionability Taxonomy (FAT) to guide template selection.
This taxonomy, which categorises features based on their level of mutability enables the use of appropriate sentence constructs for sentence planning. 
FAT is informed by findings from the user study and is used to determine the template structure for each feature, which provides alternatives for slot filling.
These options include using both query and counterfactual values, using only counterfactual values, employing ordinal adverbs or bullet points, and selecting alternative forms of action terms such as "increase" or "raise". 
Once templates are identified they are presented in order of taxonomic category using SHAP weights for within category ordering.


\subsection{Feature Actionability Taxonomy (FAT)} 
FAT was defined using a data-driven methodology that relied on examining features extracted from six datasets~\cite{Dua:2019,Quy2022} related to Fair AI. 
They span three distinct domains, with each feature analysed to determine suitable actionability categories. The resulting categories and distributions are summarised in Table \ref{tbl:taxonomy-datasets}. We observe that, while current \cfxai systems consider recipients can change all features directly, such features appear least, highlighting the need for \nxai.\label{sec:taxonomy}
\begin{table}
\begin{tabular}{llp{0.8cm}p{.5cm}p{.5cm}p{.5cm}p{.5cm}}
\hline
Domain                     & Dataset       & \#Features & \multicolumn{4}{c}{\#Features by Category} \\
                           &               &            & M.D.      & M.I.     & I.S.     & I.NS.     \\
\hline
\multirow{2}{*}{Health}    & Diabetes      & 8          & 1        & 4        & 1        & 2         \\ 
                           & Breast Cancer & 9          & 0        & 2        & 0        & 7         \\
\hline
\multirow{2}{*}{Education} & OULAD         & 8          & 0        & 2        & 4        & 2         \\ 
                           & Student (UCI) & 31         & 4        & 12       & 6        & 9    \\
\hline
\multirow{2}{*}{Finance}   & Loan Approval & 67         & 4        & 50       & 0        & 13        \\ 
                           & Income        & 8          & 0        & 2        & 1        & 5         \\
\hline
\hline
Total                      &               & 131        & 9   & 72   & 12    & 38    \\ \hline
\end{tabular}
\caption{Dataset overview with feature count and actionability category distribution based on the FAT knowledge in Figure~\ref{fig:taxonomy}. Categories: Mutable Directly (M.D.), Mutable Indirectly (M.I.), Immutable Sensitive (I.S.), and Immutable Non-sensitive (I.N.).
}
\label{tbl:taxonomy-datasets}
\end{table}


\begin{figure*}[htb]
    \centering
    \includegraphics[width=0.9\textwidth]{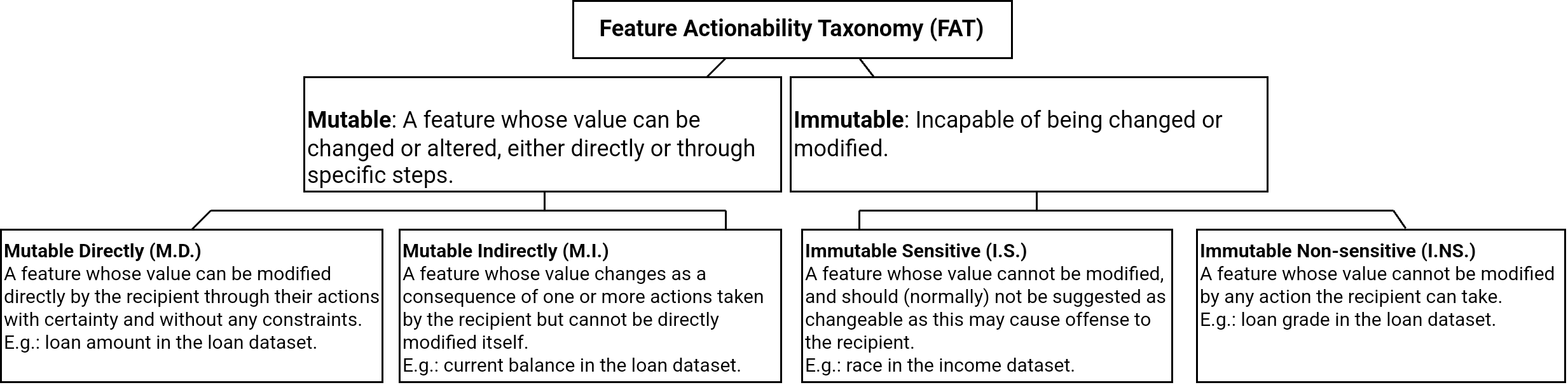}
    \caption{The Feature Actionability Taxonomy (FAT).}
    \label{fig:taxonomy}
\end{figure*}

FAT category definitions appear in Figure~\ref{fig:taxonomy}, where features are categorised into two groups: those that the recipients of counterfactuals can change through their actions (i.e., \textit{Mutable}) and those that cannot be changed (i.e., \textit{Immutable}).
This categorisation
allows \nxai to carefully consider how to present information for each category, even for immutable features. 
While recipients of explanations cannot change immutable features, classifying them in the taxonomy allows NLG systems to present these as ``factual explanations'' (in contrast to suggesting a counterfactual-driven change). 
Recognising likely ethical concerns about presenting certain features like race or ethnicity, we classify them as \textit{Immutable Sensitive} in \nxai enabling the system to consider them when generating explanation texts, 
and exposing bias with human-in-the-loop~\cite{compas}.
We further categorise those features that recipients can change into those that will be directly impacted by actions (i.e. \textit{Mutable Directly)}, 
and those that the recipient can only change by acting on another feature (\textit{Mutable Indirectly}), 
such as discretionary income, which changes by increasing salary or decreasing expenses. 
The latter category can also be useful should the XAI system provide causal knowledge. 
 \subsection{FAT Template-based NLG}
\label{sec:nlg}

\begin{figure*}[t]
    \centering
    \includegraphics[width=\textwidth]{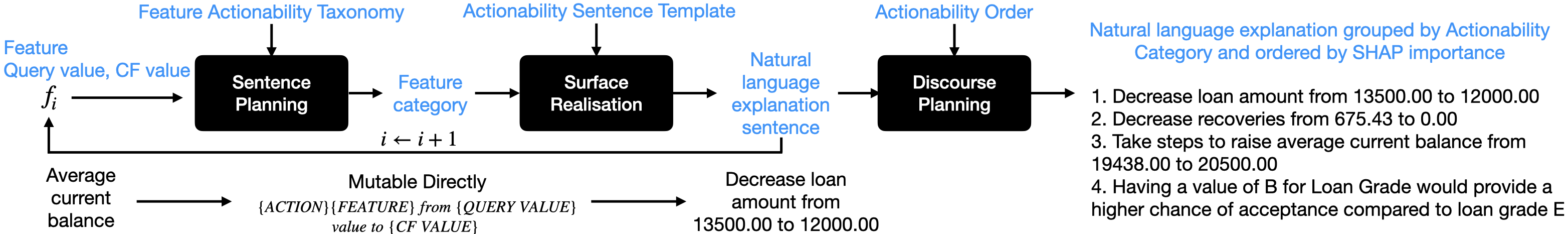}
    \caption{NLG pipeline for \nxai, using FAT and Feature Sentence Templates.}
    \label{fig:nlg}
\end{figure*}

In \nxai we adopt a three-staged template-based NLG approach of sentence planning, surface realisation and discourse planning~\cite{reiter-2007-architecture}.  
The FAT is used for sentence planning where relevant Feature Sentence Templates (see Table \ref{tbl:taxonomy-template}) are identified based on the feature's categorisation in the FAT.  
Thereafter a mapping of content and structure themes to templates guides the surface realisation step of NLG. 
This is detailed in Table \ref{tbl:taxonomy-template}, where
sentences explaining mutable features use content themes C2, C3, or C4, meaning they can opt whether to include values from the query along with the counterfactual value. Sentences explaining immutable features use theme C1 or C5. 
Further, mutable features are presented using structure theme S1 to make it easier for users to focus on them, with immutable features generally using S2.  
Additionally, we incorporate positive reinforcement language, based on insights from psychology research~\cite{burieva2020effectiveness}, into the immutable non-sensitive feature template. 
As a result, the template generates positive explanations, such as "your loan has a high chance of approval", through the use of surface realisation techniques. 
This approach effectively conveys information while maintaining a supportive tone, as opposed to negative language such as "your loan has a less chance of rejection".
Thereafter, for discourse planning, sentences are grouped by taxonomic category and sorted by a feature-based ordering (such as SHAP). 
For a given {query, counterfactual} pair, the pipeline in Figure~\ref{fig:nlg} shows the input and output for each feature at each of the three stages.
At deployment, the explanation is customised with an initial sentence specific to the dataset, including the number of actionable features, and a domain-specific epilogue, such as "Stay healthy!" for a health domain and "Good luck with your loan!" for a finance domain.

\begin{table*}[ht]
\centering
\begin{tabular}{lp{14cm}}
\hline

\multicolumn{2}{l}{\textit{Template Variables with  Synonym Examples} }\\
\hline
\multicolumn{2}{l}{\texttt{VERB=\{Take|Initiate|Undertake|Pursue|Negotiate\}}} \\
\multicolumn{2}{l}{\texttt{OBJECT=\{steps|measures|actions|\}}}  \\
\multicolumn{2}{l}{\texttt{ACTION=\{Pos: (increase|improve|raise)| Neg:(decrease|reduce)}\}} \\
\multicolumn{2}{l}{\texttt{COMPARATIVE=\{Pos: (increase|higher|better) | Neg: (decrease|lower|worse)\}} }\\
\multicolumn{2}{l}{\texttt{OUTCOME=\{undesired:(rejected|fail) | desired:(accepted|pass)\}}}  \\
\multicolumn{2}{l}{\texttt{FEATURE= feature name in dataset, QUERY\_VALUE= feature value from query, CF\_VALUE= }}\\ \multicolumn{2}{l}{\texttt{                 feature value from counterfactual, POSSESSIVE=\{Your\}}} \\
\hline
\textit{Actionability Category} & \textit{Feature Sentence Template} \\
\hline
Mutable Directly & 1. \texttt{\{ACTION\} \{FEATURE\} from \{QUERY\_VALUE\} value to \{CF\_VALUE\}} \\
& 2. \texttt{\{ACTION\} \{FEATURE\} to \{CF\_VALUE\}} \\ 
\hline
Mutable Indirectly & 1.\texttt{\{VERB\} \{OBJECT\} to \{ACTION\} \{FEATURE\} from \{QUERY\_VALUE\} to \{CF\_VALUE\}} \\
& 2.\texttt{\{VERB\} \{OBJECT\} to \{ACTION\} \{FEATURE\} to \{CF\_VALUE\}} \\
\hline
Immutable Non-sensitive & 
\texttt{Having a value of \{CF\_VALUE\} for \{FEATURE\} would provide a \{COMPARATIVE\} chance of \{DESIRED\_OUTCOME\} compared to a value of \{QUERY\_VALUE\}} \\ 
\hline
Immutable Sensitive & 
\texttt{\{POSSESSIVE\} \{FEATURE\} has contributed to \{OUTCOME\} }\\ 
\hline
\end{tabular}
\caption{Templates from mapping Content and Structure themes to Feature Actionability Taxonomy categories.}
\label{tbl:taxonomy-template}
\end{table*}

\section{Evaluating Actionability in Natural-XAI}
\label{sec:userstudy2}
This study aims to evaluate the utility of the \nxai approach to Natural-XAI using actionability knowledge. 
It was conducted across three new test domains: 
health (heart disease dataset), 
finance (credit risk - kaggle)
and education (student performance - kaggle).
Manual instantiation of the taxonomy for each dataset is summarised in Table~\ref{tbl:taxonomy-datasets-val}. 
Using \cfxai system (DICE), 
six scenarios were devised, two per domain, and a comparative analysis was conducted with two cohorts: Cohort1 using a baseline \basexai and Cohort2 using \nxai. 
We improved the formatting of \bxai based on previous findings, resulting in the creation of \basexai. 
The key difference between the two is in the handling of mutable features. 
\nxai generates factual explanations for features in categories I.N and I.NS, 
e.g., "$\ldots$ your parental level of education is a contributing factor to the risk of obtaining an overall credit risk score of below average".
In contrast \basexai treats all features as  actionable  e.g., " $\dots$ change your parental level of education to bachelor's degree". 
Further examples for each domain are provided in the section \ref{sec:src}.

\begin{table}
\begin{tabular}{llp{0.8cm}p{.5cm}p{.5cm}p{.5cm}p{.5cm}}
\hline
Domain                     & Dataset       & \#Features & \multicolumn{4}{c}{\#Features by Category} \\
                           &               &            & M.D.      & M.I.     & I.S.     & I.NS.     \\
\hline
Health                     & Heart      & 13         & 0        & 8        & 2        & 3        
       \\
Education     
                           & Student (Kaggle) & 8         & 5        & 0       & 3       & 0    \\
Finance                    & credit & 11         & 2        & 1       & 7        & 1            \\
\hline
Total                      &               & 32        & 7   & 9   & 12    & 4    \\ \hline
\end{tabular}
\caption{Overview of test datasets in each domain, displaying the number of features and the actionability category distribution.}
\label{tbl:taxonomy-datasets-val}

\end{table}

\subsection{User Study Protocol}
Our study consisted of a non-randomised between-subjects design with three domains, each with two scenarios. A total of 60 participants were prescreened and assigned to one of two cohorts based on their domain knowledge alignment, with 20 participants allocated to each domain (and divided into 10 per cohort). 
Cohort1 was presented with the \basexai explanations for the two scenarios in their respective domain, while Cohort2 was presented with explanations generated using the \nxai pipeline. Each cohort completed the same two scenarios in their respective domain. 
The between-subjects design allowed us to compare the effectiveness of the two types of explanations across different domains and independent cohorts while controlling for individual differences between participants.

After recruiting participants for our study, we carefully screened their responses to ensure that only reliable and qualified participants were included. Upon close examination of the responses, we discarded six responses that failed the attention test, resulting in a final sample of 54 participants. 
Our participant pool consisted of native English speakers from the USA, UK, Ireland, Australia, Canada, and New Zealand, with an equal distribution of participants based on sex. 
Additionally, we prescreened participants on age, ranging from 18 to 74 years old for the health and education domains, and from 28 to 74 years old for the finance domain. 
We also prescreened them based on relevant domain expertise.

Given the unique nature of each domain, we conducted separate user studies for each domain, with different groups of participants completing the scenarios for each domain. 
After being presented with the two scenarios, participants were asked to answer 4 questions related to the explanations they received, evaluating them based on the following criteria: Articulation, Acceptability, Feasibility, and Sensitivity. Participants rated their response to each question using a 5-point Likert scale~(1-5) and were also asked to provide their rationale for selecting a specific rating.
The results from each domain were analysed separately, allowing us to draw distinct conclusions and insights for each domain.
The user studies were conducted on the Prolific platform, a reputable source for obtaining high-quality data from diverse participant samples using the prolific platform. 

\subsection{User Study Outcomes}
\begin{figure*}[htb]
     \centering
     
     \begin{subfigure}[b]{0.325\textwidth}
         \centering
         \includegraphics[width=\textwidth]{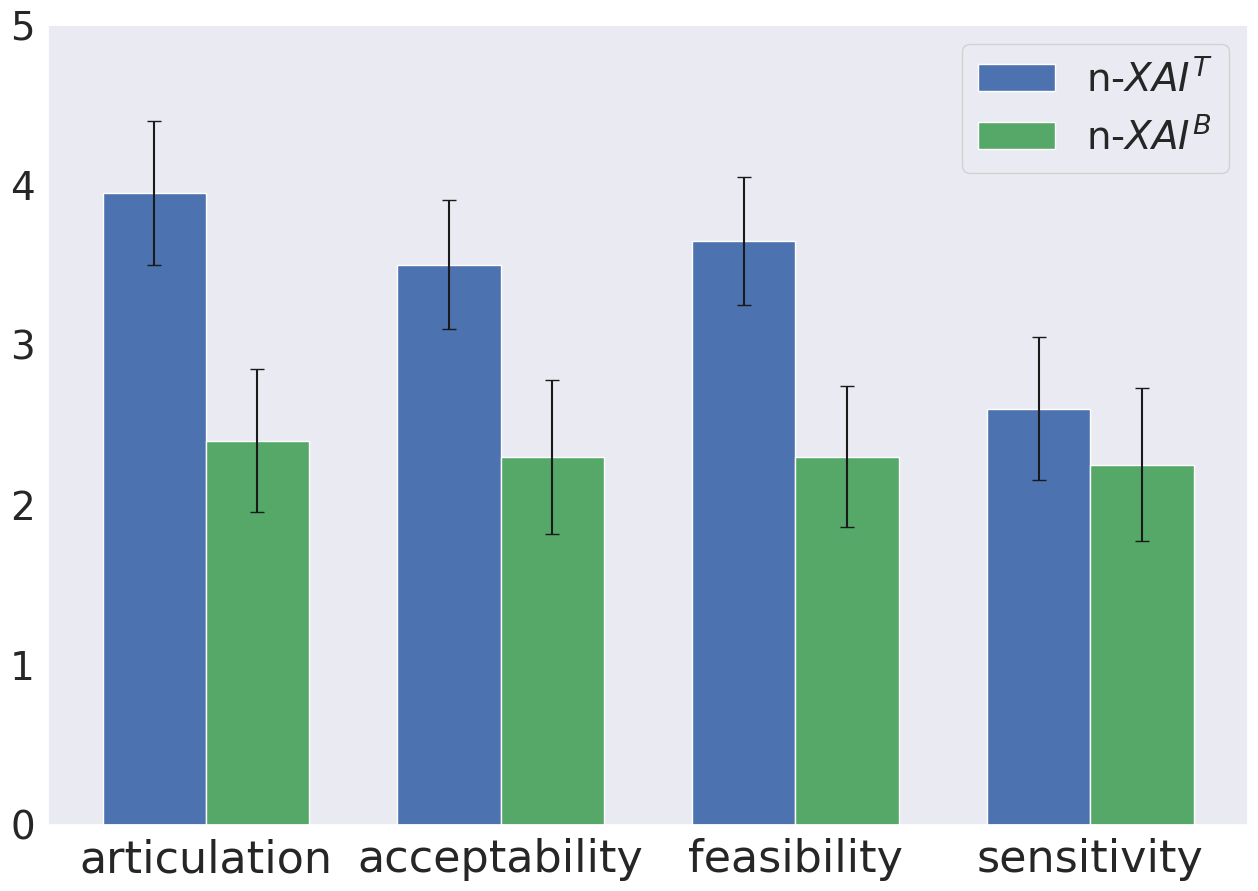}
         \caption{Healthcare Domain}
         \label{fig:us2-health}
     \end{subfigure}
     \hfill
     \begin{subfigure}[b]{0.325\textwidth}
         \centering
         \includegraphics[width=\textwidth]{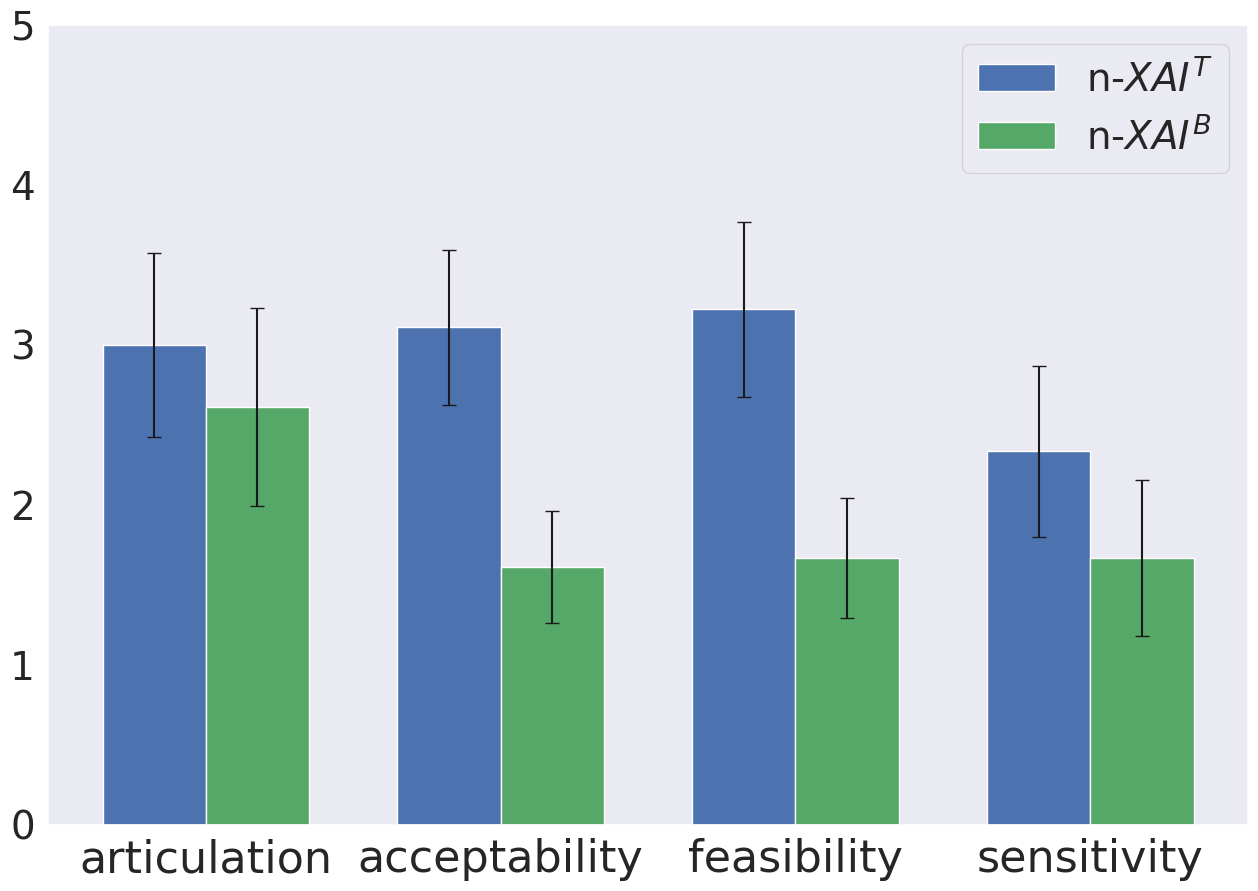}
         \caption{Education Domain}
         \label{fig:us2-edu}
     \end{subfigure}
     \hfill
     \begin{subfigure}[b]{0.325\textwidth}
         \centering
         \includegraphics[width=\textwidth]{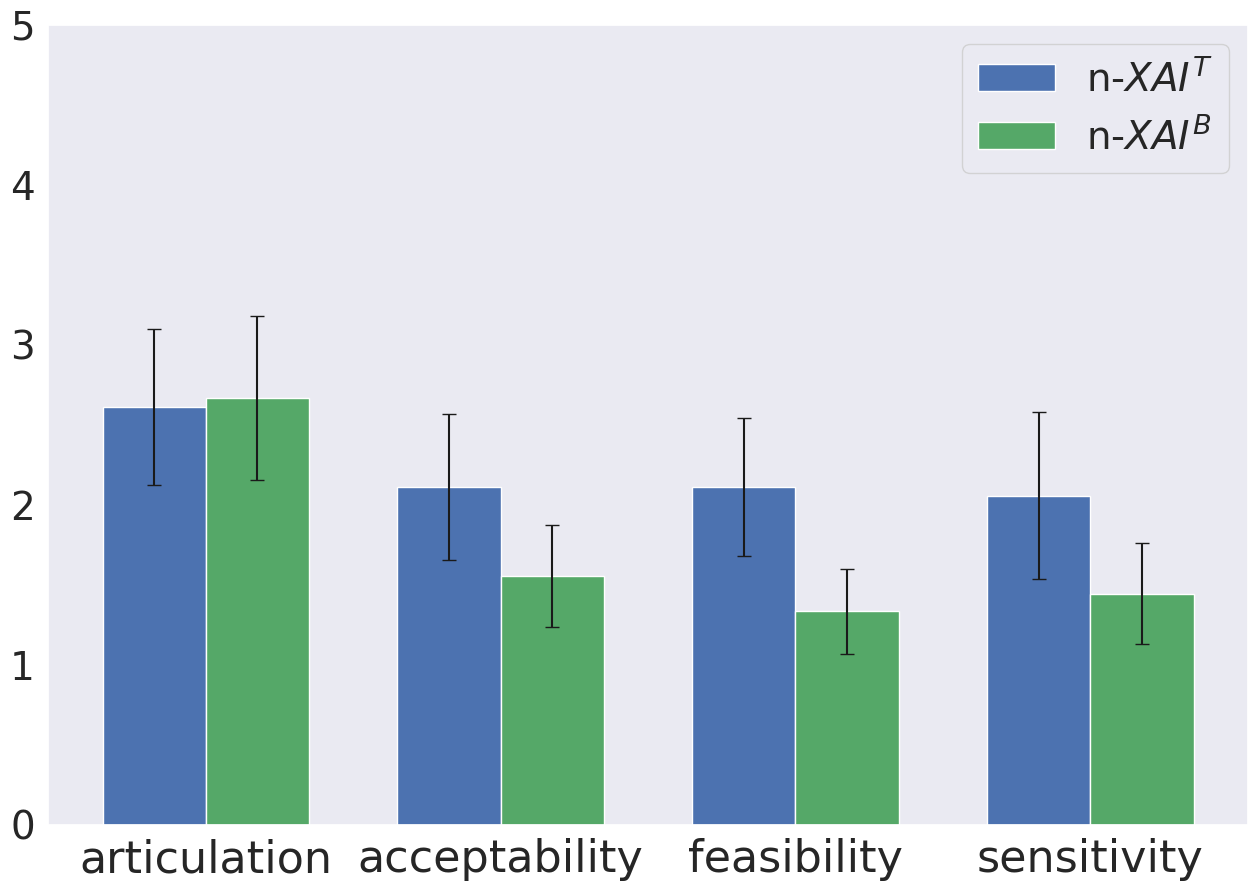}
         \caption{Finance Domain}
         \label{fig:us2-fin}
     \end{subfigure}
\caption{Comparison of participant responses between \nxai and \basexai across all domains.}
\label{fig:us2}
\end{figure*}

\begin{table*}[!htb]
\small
\centering
\begin{tabular}{lll}
\hline
\textit{Theme} & \textit{Domain(s)} & \textit{Example of user suggestions} \\ \hline
Feasibility Knowledge & \begin{tabular}[c]{@{}l@{}}H, E, F\end{tabular} & 
\begin{tabular}[c]{@{}p{12.5cm}@{}} 
"The 4 \textbf{steps} to take are \textbf{better at indicating what can be done} to offer practical advice" - H2\\ "The patient should be given \textbf{suggestions} for how to decrease their resting BP
" - H2 \end{tabular} \\ \hline
\begin{tabular}[c]{@{}l@{}}Handling Protected\\  Features\end{tabular} & \begin{tabular}[c]{@{}l@{}} \\ H, E, F\end{tabular} & 
\begin{tabular}[c]{@{}p{12cm}@{}}
"The suggestion to reduce her age and subtype of thalassemia \textbf{are essentially discriminatory} and \textbf{should not be included}." - H1
\\"It is \textbf{highly unethical} to suggest someone changes their sex" - E1\\ 

"Statements were \textbf{factual \& not offensive} although they could be interpreted this way" - E2,\\ "Not taking into account work history of 5 years \& consistent employment is \textbf{unethical}" - F2 \end{tabular}  \\ \hline
Hybrid Explanations & \begin{tabular}[c]{@{}l@{}}H, E\end{tabular} & 
\begin{tabular}[c]{@{}p{12cm}@{}}"...\textbf{reasons} for these need to be \textbf{clear}, and the \textbf{means with which to achieve} them " - F2\\ 
"Suggest \textbf{reasons} as to why he may be behind but these are not evidenced" - E2 \end{tabular} \\ \hline
\begin{tabular}[c]{@{}l@{}}Personalised Revision\\of Counterfactuals\end{tabular} & F & 
\begin{tabular}[c]{@{}p{12cm}@{}}
"...it would be \textbf{hard to change} ownership to rent as you  would need to sell your house" -  F2\\ "Considering her age and income /risk again \textbf{not sure this is suitable} for subjects unless they work in finance and \textbf{understand these terms}" - F2 \end{tabular} \\ \hline
Feasibility Impact & F & \begin{tabular}[c]{@{}p{12cm}@{}}
"The potential \textbf{negative outcomes are not provided} to the user by the AI,
meaning that harm may come to them by following advice that they do not understand" - F1\\ 
"...limit the suggestions to things that are \textbf{actually possible} for a person to do" - F1\end{tabular} \\ \hline
Structure and Context & \begin{tabular}[c]{@{}l@{}} H, E, F\end{tabular} & \begin{tabular}[c]{@{}p{12cm}@{}} 
"Tense \& use of \textbf{capitals} \& upper case letters prevent advice from being well articulated" - F2\\
"Suggestion is too wordy - E2"
\end{tabular} \\ 
\hline
\end{tabular}
\caption{User response analysis: explanation improvement themes (column 1), domain-wise theme coverage (column 2), and mix of user quotes from the domains, Healthcare (H), Education (E) \& Finance (F) and Cohort (1 or 2) (column 3).}
\label{tab:us2thematic}
\end{table*}

Figure \ref{fig:us2} presents the results for each domain on the 4 criteria with 95\% confidence bars. 
The Health domain received the highest ratings, followed by Education and Finance.
We conducted a Shapiro-Wilk test to determine the normal distribution of both cohorts for each domain, followed by Levene's test to assess variance equality. 
Based on the outcomes of these tests, we used the Wilcoxen Signed Rank test to evaluate the significance of each domain's cohorts for the 4 criteria.
Our analysis revealed significant improvements in the Feasibility ratings for explanations across all domains. 
Incorporating actionability knowledge has also resulted in significantly Acceptable recommendations in the Education and Health domains. Although \nxai outperformed \basexai in terms of Articulation and Sensitivity in all domains, the differences were not statistically significant. 



All textual responses were thematically analysed to identify areas for improvement in the explanations provided. Table~\ref{tab:us2thematic} shows the identified themes and provides example participant responses from both cohorts.
Further analysis of these responses, indicates that users' subjective perceptions of their expectations significantly influenced their feedback, with this trend being particularly noticeable in the Finance domain (acceptability criteria).
For instance, those who saw \nxai explanations wanted more detailed strategies and explanations on how to achieve the suggested changes, which would require extensive domain knowledge beyond the scope of this study~(see quotes in themes Feasibility Knowledge and Hybrid Explanations for example). 
Conversely, those who saw baseline explanations found them to be less acceptable and feasible due to their perceived lack of rationality and real-world applicability (theme Handling Protected Features).

Regarding Sensitivity and articulation, in both \nxai and \basexai, concerns remained about the need for personalisation and ethically considerate explanations.
For instance, recommending actions related to changing ownership to rent (from a mortgage) applies only to those that don't already own a house; but to support this level of inference requires the XAI system to have access to the user's background, which may not always be practical. 
These observations emphasise the user-dependent nature of applying taxonomy definitions. To address this in deployed systems, an interactive iterative process with the user is necessary, enabling adaptive feature actionability categories based on individual circumstances. This approach aligns with prior research advocating for interactive and personalised XAI systems~\cite{wijekoon_iccbr23}.
Many responses about the structure emphasised the need for enhanced articulation. 
Suggestions included using intuitive names, and employing appropriate capitalisation, and adopting understandable units of measurement.
Interviews to understand expectations were desirable but not possible due to the online evaluation.
Overall we found common themes related to wanting more guidance on how to achieve suggested changes across all domains. This was especially true of Finance. The use of factual explanation style for sensitive features and counterfactual action recommendations for others in \nxai was liked by Cohort2 participants, especially in Health and Education (see Hybrid theme).

 \section{Conclusion}
This paper presents \nxai, a Natural-XAI approach that enhances natural language counterfactual explanations with actionability knowledge, resulting in better results for articulation, sensitivity, feasibility, and actionability criteria. 
Guided by an actionability taxonomy, and feature attribution weights, \nxai selects feature-based sentence-level templates to generate natural explanations. 
Results from a user study (n=60) provide useful guidelines for counterfactual XAI platforms to enhance the feasibility of recommended actions, including incorporating mixed XAI strategies (factual and counterfactual), use of domain knowledge to guide users on how to implement recommended changes and personalising actionability categories to individual preferences.
The taxonomy is open-sourced for community contributions in new domains, and future work includes extending the taxonomy, simplifying structures for accessibility, and tailoring language templates to user personas.

\section*{Acknowledgements}
This research was partially funded by the iSee project (https://isee4xai.com/). iSee is an EU CHIST-ERA project which received funding for the UK from EPSRC under grant number EP/V061755/1. We are grateful to the University Research Ethics Committee for approving the study protocol. Special thanks to all our participants, who were fully informed about their right to withdraw and the handling and storage of their data. Finally, we wish to thank all our anonymous reviewers for their valuable feedback that helped to improve the paper. 

\section*{Resources}
\label{sec:src}
All code and materials from this paper are available from \url{http://github.com/pedramsalimi/NLGXAI}.


\end{document}